\documentclass{article}
\usepackage{spconf,amsmath,graphicx,tabularx}
\usepackage{booktabs}
\usepackage{float}
\usepackage{caption}
\usepackage{wrapfig}
\usepackage{verbatim}
\usepackage{graphicx}
\usepackage{pdfpages}

\usepackage[colorlinks=true, allcolors=blue]{hyperref}

\usepackage{amssymb}

\title{GPT-4 as Evaluator: Evaluating Large Language Models on Pest Management in Agriculture}
\name{Shanglong Yang, Zhipeng Yuan, Shunbao Li, Ruoling Peng, Kang Liu, Po Yang*}
\address{}
\begin{document}

\maketitle
\thispagestyle{plain}
\pagestyle{plain}
\setcounter{page}{1}
\pagenumbering{arabic}

\begin{abstract}

In the rapidly evolving field of artificial intelligence (AI), the application of large language models (LLMs) in agriculture, particularly in pest management, remains nascent. We aimed to prove the feasibility by evaluating the content of the pest management advice generated by LLMs, including the Generative Pre-trained Transformer (GPT) series from OpenAI and the FLAN series from Google. Considering the context-specific properties of agricultural advice, automatically measuring or quantifying the quality of text generated by LLMs becomes a significant challenge. We proposed an innovative approach, using GPT-4 as an evaluator, to score the generated content on Coherence, Logical Consistency, Fluency, Relevance, Comprehensibility, and Exhaustiveness. Additionally, we integrated an expert system based on crop threshold data as a baseline to obtain scores for Factual Accuracy on whether pests found in crop fields should take management action. Each model's score was weighted by percentage to obtain a final score. The results showed that GPT-3.4 and GPT-4 outperform the FLAN models in most evaluation categories. Furthermore, the use of instruction-based prompting containing domain-specific knowledge proved the feasibility of LLMs as an effective tool in agriculture, with an accuracy rate of 72\%, demonstrating LLMs' effectiveness in providing pest management suggestions.
\end{abstract}
%

\section{Introduction}
\label{sec:intro}

Language models (LMs), as computer algorithms or systems, are capable of understanding and generating human language, contributing a core component of the field of natural language processing (NLP) \cite{3}. Language models are trained on a vast corpus of text data \cite{6}, enabling the model to capture word order or contextual associations, which allows the model to predict the next word or a sequence of words based on a particular probability distribution given an input \cite{6,7,8}. LLMs are sophisticated LMs with a considerably larger scale, encompassing billions or hundreds of billions of parameters, and are typically founded upon deep learning methodologies \cite{3}. In contrast to standard LMs, LLMs necessitate massive data for training, thereby enabling LLMs with a broad expanse of knowledge and generalization capabilities. LLMs exhibit enhanced adaptability to a diverse range of tasks and domains \cite{4,5}. Large, pre-trained language models (PLMs) like BERT (Bidirectional Encoder Representations from Transformers) and GPT have significantly altered the NLP landscape, delivering state-of-the-art results across various tasks \cite{49}. Traditional NLP methods require handcrafted features and task-specific training, whereas PLMs use a generic latent feature representation learned from extensive training on a wide range of texts adapted for specific NLP tasks \cite{49}.

LLMs such as GPT-3.5 and GPT-4 have demonstrated remarkable capabilities as general-purpose computational tools, conditioned by natural language instructions. The efficacy of these models in task performance is substantially contingent upon the quality of prompts used to guide them. Notably, most effective prompts are crafted manually by humans \cite{05}. Prompt engineering emerges as a pivotal area within AI, dedicated to optimising prompts to proficiently direct AI models, particularly those grounded in machine learning and NLP. The emerging research domain includes the design, refinement, and implementation of prompts or instructions that steer the output of LLMs, facilitating the completion of diverse tasks.

Generally, LLMs are pre-trained on a massive corpus of unlabeled data to capture a broad understanding of language and knowledge. Followed by small fine-tuning, LLMs are adapted to task-specific datasets to particular applications of interest \cite{25}. Consequently, identifying appropriate evaluation metrics for LLMs across diverse domains has emerged as a novel and significant research theme. Due to the efficiency in understanding and generating human language, LLMs have been applied across various domains, including finance, medicine and education. However, their adoption in agriculture has been limited, constrained by the field's specialized nature and the paucity of research exploring their potential in this area.

The main contributions of our paper can be summarized as follows:
\begin{enumerate}

    \item \textbf{Feasibility Study of LLMs for Pest Management Advice Generation in Agriculture:} We demonstrate the viability of LLMs in the agricultural pest management domain.

    \item \textbf{Innovative Evaluation Methodology:} We introduce a novel approach using GPT-4 for multi-dimensional assessment of generated pest management suggestions.

    \item \textbf{Effective Application of Instruction-Based Prompting Techniques:} Our findings highlight a 72\% accuracy in LLM-driven pest management decisions through instruction-based prompting that incorporates domain-specific knowledge.

    \item \textbf{Nuanced Differences Between GPT-3.5 and GPT-4:} Our research uncovers subtle differences between GPT-3.5 and GPT-4 in decision-making on pest management, emphasizing the importance of model selection in agricultural contexts.

\end{enumerate}

\section{Related Work}
\label{sec:related work}

\subsection{Application of LLMs}
``FinBERT'' is a LM tailored in financial domain, a variant of the BERT model where lies in the specialized pre-training on financial texts, enabling the adaptability to handle the distinctive language and expressions prevalent in the financial sector. ``FinBERT'' has been applied for financial text mining \cite{50}, financial sentiment analysis \cite{51}, and financial communications \cite{52}. However, the specialization of ``FinBERT'' is limit on effectiveness in domains outside of finance as the model's performance is highly dependent on the quality and representativeness of the financial corpus used for training \cite{50,51,52}. Beyond ``FinBERT'', Xiao-Yang et al. \cite{53} have introduced ``FinGPT'', a novel model based on the transformer architecture, aimed at enhancing the applicability of LLMs in the financial domain. ``FinGPT'' addresses the limitations in data acquisition and processing faced by traditional financial LLMs by automating the collection of real-time financial data from the Internet. In evaluating LLMs in educational domains, Kung et al. \cite{032} demonstrated that ChatGPT could achieve scores at or near the passing threshold for all three components of the United States Medical Licensing Exam without specific training or reinforcement, underscoring the potential of LLMs to support medical education and possibly influence clinical decision-making processes. Similarly, Thirunavukarasu et al. \cite{033} discussed the use of LLMs in healthcare, which covered development and applications in clinics. The review guides clinicians on using LLM technology for patient and practitioner benefits.

In agriculture, Dr Som \cite{56} explored the potential applications of OpenAI's LLM, ChatGPT. Specifically, the paper discusses using ChatGPT across various agricultural tasks, including crop forecasting, soil analysis, crop disease and pest identification. Dr Som highlights that ChatGPT exhibits professional competence in analyzing agricultural data to generate accurate and timely reports, alerts, and insights, facilitate informed decision-making, and enhance customer service. However, it is noted that ChatGPT's predictions' accuracy relies heavily on input data quality. Inaccurate, biased, or incomplete data can significantly impact the model's outputs. Moreover, AI systems like ChatGPT can assist decision-making but are not a substitute for human intuition and experience in complex agricultural environments \cite{56}. Besides, Silva et al. \cite{54} evaluate the capability of LLMs, including GPT-4, GPT-3.5, and Llama2, in responding to agriculturally-related queries. The queries were sourced from agricultural examinations and datasets from the United States, Brazil, and India. The study assessed the accuracy of answers produced by LLMs, the effectiveness of retrieval-augmented generation (RAG) and ensemble refinement (ER) techniques, and the comparative performance against human respondents. Silva et al. \cite{54} discovered that in various tasks, GPT-4 performed better than GPT-3.5 and Llama2, achieving an impressive 93\% accuracy rate in the certified crop adviser (CCA) exam. Additionally, in the study by Jiajun et al. \cite{55}, the application of LLMs, particularly GPT-4, in agriculture for pest and disease diagnosis is explored. Jiajun \cite{55} introduces a novel approach that combines the deep logical reasoning capabilities of GPT-4 with the visual comprehension abilities of the You Only Look Once (YOLO) network. The paper evaluates the YOLO-PC, a new lightweight variant of YOLO, using metrics such as accuracy rate (94.5\%) and reasoning accuracy (90\% for agricultural diagnostic reports), assessing the quality of model-generated text in correlation with the recognized information \cite{55}.

\subsection{Prompt \& Prompt Engineering}
\label{sec:prompt}

Prompts are a mechanism for interaction with large language models (LLMs) to accomplish specific tasks \cite{01}. Prompts act as essentially instructions directed towards LLMs, comprise the input provided by users and guide the model to generate answers for the response \cite{02}. The nature of the inputs are vary, encompassing explanations, queries, or any other form of input, contingent upon the intended application of the model \cite{01}. In contrast to traditional supervised learning, where models are trained to predict output from input using a probability distribution, prompt-based learning operates on LLMs that directly model textual probabilities. Prompt-based learning involves modifying the original input into a text string prompt with unfilled slots using templates. Subsequently, prompts are populated using the probabilistic capabilities of the LLM to generate the final string \cite{03}. Essentially, prompt engineering represents a practice of engaging effectively with AI systems to optimise the utility \cite{06}. In addition, prompt engineering has been applied in various domains such as medical \cite{06,09,010}, generative art \cite{011}, multilingual legal judgment prediction \cite{08}, and the extraction of accurate materials data \cite{07}.

As delineated in the ``Prompt Engineering Guide \cite{04}'', constructing an effective prompt can involve integrating four elements or a combination: Instruction, Context, Input Data and Output Indicator. Instruction refers to a specific task or directive to guide the model to perform a designated operation. Context encompasses providing supplementary information or background, instrumental in steering the model towards more accurate responses. Input Data pertains to the specific question or input content the model solicits to respond to. Lastly, the Output Indicator concerns the desired type or format of the model's output.

The iterative development process also outlined four prompt guidelines \cite{012}:

\begin{itemize}
  \item Be clear and specific: The prompts should be unambiguous and detailed enough to guide the model precisely towards the intended task or output.
  \item Analyze why the result does not give the desired output: If the output from the prompt does not meet expectations, it is crucial to analyze the reasons behind the discrepancy.
  \item Refine the idea and the prompt: Based on the analysis, adjustments should be made to both the underlying idea and the wording or structure of the prompt to improve results.
  \item Repeat: The process is not linear but cyclical, after refining, the new prompt is tested, and the cycle of analysis and refinement continues until the desired outcome is achieved.
\end{itemize}

\section{Experiment Design}
\subsection{Experiment Models}
\label{sec:models}
This section provides an overview of the two LLMs evaluated in the experiment: Section \ref{GPT} covers the GPT series from OpenAI, specifically GPT-3.5 and GPT-4, while Section \ref{FLAN-T5} describes the FLAN-T5 model developed by Google.

\subsubsection{GPT}
\label{GPT}
The transformer architecture, proposed by Vaswani et al. in the paper ``Attention is All You Need'' \cite{16}, became the cornerstone for the GPT \cite{17,18}. The OpenAI GPT model, introduced in the paper "Improving Language Understanding by Generative Pre-Training" \cite{024}, undergoes pre-training through language modelling on a substantial dataset to capture long-range dependencies within the text. Due to the GPT model's advanced capability to understand and generate human-like text \cite{19}, it becomes an ideal choice for exploring complex agriculture tasks and serves as the experimental model. Specifically, the GPT-3.5 (Model: `gpt-3.5-turbo-0125') and GPT-4 (Model: `gpt-4-1106-preview') models were used in the experiments.

GPT-3.5 and GPT-4 are successive generations of artificial intelligence language models developed by OpenAI. The GPT-3.5 model is proficient in understanding and generating natural language or code and has been optimized specifically for chat-based interactions through the Chat Completion API. However, it remains applicable to non-chat tasks. The GPT-4 model, as a large multi-modal model, exhibits a broader comprehension of general knowledge and reasoning capabilities, enabling it to solve complex problems with greater accuracy compared to GPT-3.5 and its predecessors \cite{025}.

\subsubsection{FLAN-T5}
\label{FLAN-T5}
The T5 model significantly advances natural language processing through its novel unified framework. T5 converts all language problems into a text-to-text format, facilitating extensive exploration of transfer learning techniques. Employing a combination of supervised and self-supervised training methods, including a novel use of corrupted tokens for pre-training, T5 sets new benchmarks across a range of NLP tasks by leveraging its encoder-decoder architecture and the extensive ``Colossal Clean Crawled Corpus" \cite{027}. FLAN-T5 is an evolution of the original T5 model, which was fine-tuned on over a thousand additional tasks and expanded language coverage. FLAN-T5 significantly enhances performance and versatility, even in few-shot scenarios, achieving state-of-the-art results on various benchmarks \cite{028}.

The FLAN-T5 model is available in various sizes, including Small, Base, Large, XL, and XXL, with the XXL version being the largest, encompassing 11 billion parameters. Unlike the GPT model, FLAN models require downloading the checkpoints locally to generate the response. Given the considerations for computational speed and memory constraints, the FLAN-T5-XL variant is selected as a more practical option for experimental use containing 3 billion parameters. The pre-trained model ``google/flan-t5-xl'' weights and configuration are loaded using the \verb|transformers| library. The weights and configurations are based on the previously saved checkpoint containing all the model parameters.

\subsection{Baselines}
This section elucidates the methodology for generating labelled samples used to construct pest scenarios based on the expert system. To assess the ability of LLMs to determine whether specific pest scenarios necessitate action, a baseline of labelled samples is essential. Section \ref{es} delineates the composition of the expert system, including four data files, while Section \ref{ges} elaborates on the process of generating labelled samples from the expert system's data for the construction of pest scenarios.

\subsubsection{Expert System}
\label{es}

As the baseline for this experiment, an Expert System is used to evaluate the Factual Accuracy of three different Large Language Models (LLMs) below on whether pests found in the crop fields necessitate management actions. The Expert System comprises four datasets extracted from the AHDB's Encyclopaedia of Pests and Natural Enemies in Field Crops \cite{023}. These datasets include two in structured data `JSON' format: `pest\_to\_affected\_crop.json' and `pest\_to\_threshold.json' and two in `XLSX' format `thresholds\_database.xlsx' and `pest\_to\_management.xlsx'.

\begin{itemize}
    \item \textbf{File `pest\_to\_affected\_crop.json'} summarises various pests and the crops. It lists different types of pests where each pest is associated with one or more crops.
    \item \textbf{File `pest\_to\_threshold.json'} provides information on the thresholds for pests, specifying when action should be taken to manage them. Each entry includes the pest name and a threshold description, which details the criteria for deciding when to take action, such as the temperature, location, plant stages, pest density levels and the extent of crop damage.
     \item \textbf{File `thresholds\_database.xlsx'} features a first column listing the names of pests, with other columns containing threshold information extracted from the file `pest\_to\_threshold.json'. The threshold includes pest density metrics such as `per square meter', `per plant', `leaf area eaten', `per trap', `of petioles damaged', `of plants are infested', `of tillers infested', `per pheromone trap', `per ear', `per trap on two consecutive occasions', `per yellow sticky trap', and `per gram of soil'.
    \item \textbf{File `pest\_to\_management.xlsx'} has two columns, the first listing the names of pests and the second detailing management suggestions for Non-chemical control solutions that meet the criteria for affected plants and thresholds achieved. Notably, the Expert System is designed to output Non-chemical control solutions only when all specified conditions are met, defined as action is necessitated.
\end{itemize}

As the benchmark for Factual Accuracy, the Expert System is not engaged in evaluating the accuracy, being designated as unequivocally 100\% accurate. Only the outputs, Non-chemical control solutions, from expert systems are subject to evaluation by GPT-4, focusing on Coherence, Logical Consistency, Fluency, Relevance, Comprehensibility, and Exhaustiveness.

\subsubsection{Generation of Input Samples from Expert System}
\label{ges}

\begin{figure}[htbp]
    \centering
    \includegraphics[page=1, width=\linewidth, height=.5\textheight, keepaspectratio]{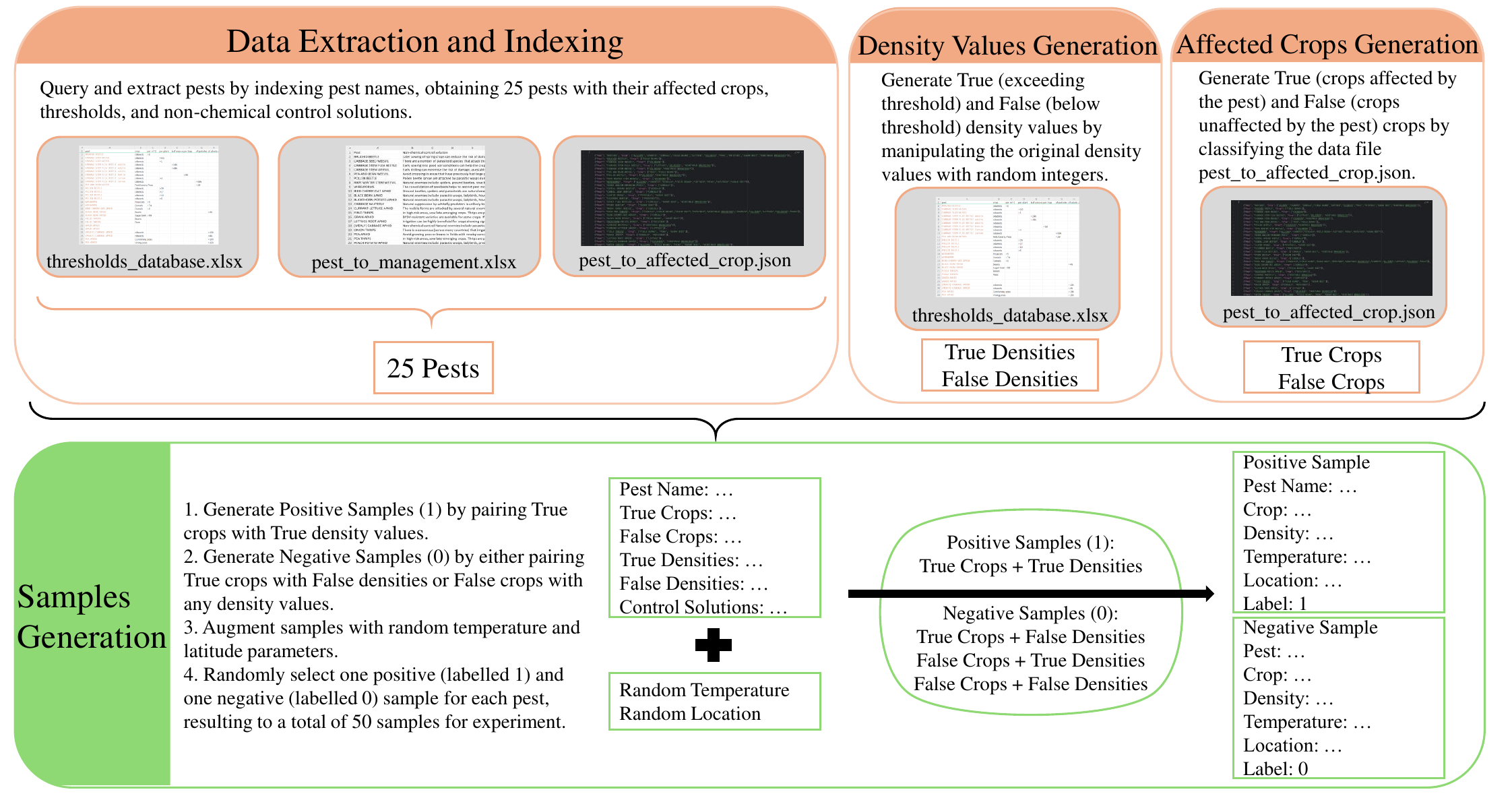}
    \caption{Generation of Input Samples from Expert System. This image outlines a process for generating labelled pest samples, detailing steps from data extraction and indexing of pests, through generating true and false density values and crops, to the creation of positive and negative samples for experiment.}
    \label{fig:ges}
\end{figure}

Figure \ref{fig:ges} shows the process for generating labelled pest samples. Files `pest\_to\_affected\_crop.json', `thresholds\_database.xlsx', and `pest\_to\_management.xlsx' are used for the generation of labelled pest samples, serving as inputs for constructing prompts for LLMs. Although the data across these files are indexed by pest name, variations exist in the pests included due to the differing extraction methods employed from the AHDB database \cite{023}. By querying pests of the same species, 25 types of pests, along with their affected crops, thresholds and Non-chemical control solutions, have been extracted.

The `generate\_densities' function provides a mechanism for determining `true' and `false' density values. By iterating through a list of density-related columns in file `thresholds\_database.xlsx', the function searches for non-null entries that signify recorded density thresholds. When encountering a valid density value, the function performs a series of operations to cleanse and standardize the data, including removing percentage symbols or relational operators. Subsequently, the numerical density value is manipulated to generate a series of `true' densities, inflating the original value by adding a random integer ranging from 1 to 10 to simulate density conditions exceeding the threshold for pest management action. Conversely, `false' densities are generated by subtracting a random integer from the original value, ensuring the resultant value does not fall below zero. These reduced values represent conditions below the pest management threshold, indicating no action is needed. The generated `true' and `false' density values are then appended with the original measurement metric (e.g., `per plant', `per square meter') and a percentage symbol if the original value was expressed as a percentage.

The core of the sample generation process is to create the combinations that represent an action that is needed (labelled as `1') and not needed (labelled as `0') for various pest conditions. This bifurcation is achieved by deliberately pairing crops and pest density values under varying conditions. Positive samples are formulated by coupling `true' crops (crops affected by the pest) with `true' density values. Conversely, negative samples emerge from the strategies: pairing `true' crops with `false' densities and `false' crops (crops unaffected by the pest) with either `true' or `false' densities. These combinations are augmented with randomly generated temperature and latitude location parameters to diversify the dataset further.

Considering computational resources and experimental costs, also ensuring an even distribution of positive and negative samples, one positive sample (labelled as `1') and one negative sample (labelled as `0') are randomly selected for each of the 25 pest types. Eventually, a total of 50 samples are generated for experimentation. The samples are indexed in pest names, with other columns containing crops, pest density, temperature and location. Among these, crops and pest density determine the label as 1 or 0, whereas temperature and location only enrich the scene and do not affect the label.

\subsection{Experiment Prompting}
This section lists the prompts constructed using different techniques in the experiment. Four prompt techniques: zero-shot prompting described in Section \ref{zs}, few-shot prompting in Section \ref{fs}, instruction-based prompting in Section \ref{ib}, and self-consistency prompting in Section \ref{ss}, incorporate samples of pest scenarios generated in Section \ref{ges} into prompts. These prompts serve as inputs for LLMs to generate responses, which are then evaluated.

\subsubsection{Zero-shot Prompting}
\label{zs}
Zero-shot prompting refers to providing instructions or requests to LLMs without needing prior examples or contextual information. Zero-shot prompting necessitates the ability of the model to comprehend and respond to tasks or queries not directly encountered before \cite{015}. The model relies on the extensive knowledge and understanding acquired during the training phase for zero-shot prompting. For instance, when posed with a question that has not previously been addressed, the model can still understand and attempt to provide an answer \cite{016}.

For zero-shot prompting, 50 input samples are iteratively filled into the following prompt template via a loop: \emph{I discovered \{Pest\} in my \{Crop\}, with a density of \{Density\}. The temperature was \{Temperature\}, and the location was at \{Location\}. Could you please provide some control and management suggestions?} This prompt is then input into a GPT or FLAN model.

\subsubsection{Few-shot Prompting}
\label{fs}
In contrast to zero-shot prompting, few-shot prompting provides relevant examples to guide the model to understand and execute a task. Few-shot prompting can be employed to facilitate in-context learning, where demonstrations in the prompt guide the model towards enhanced performance \cite{017}. According to Min et al. \cite{018}, in the context of few-shot learning, both the label space and the distribution of the input text defined by the demonstrations are crucial for performance, irrespective of the accuracy of individual labels. Additionally, the format of the demonstrations, including random labels, significantly influences effectiveness, which is better than not using any labels.

The core of the few-shot learning approach is encapsulated within a \verb|create_prompt| function. The function filters 50 input samples to select only samples with a label of `1' and a pest different from the current input pest. It randomly selects three samples and constructs a few-shot prompt containing questions and answers. Each question is formulated from the pest, crop, density, temperature, and location of the selected input samples, same as the zero-shot prompting template, followed by the respective Non-chemical control solutions from file `pest\_to\_management.xlsx'. Finally, the prompt adds a new question using the current input sample without providing an answer. The template of the few-shot prompt is shown below: \emph{\newline \textbf{Question:} I discovered \{Pest 1\} in my \{Crop 1\}, with a density of \{Density 1\}. The temperature was \{Temperature 1\}, and the location was at \{Location 1\}. Could you please provide some control and management suggestions? \newline \textbf{Answer:} \{Non-chemical control solutions for Pest 1\} \newline \textbf{Question:} I discovered \{Pest 2\} in my \{Crop 2\}, with a density of \{Density 2\}. The temperature was \{Temperature 2\}, and the location was at \{Location 2\}. Could you please provide some control and management suggestions? \newline \textbf{Answer:} \{Non-chemical control solutions for Pest 2\} \newline \textbf{Question:} I discovered \{Pest 3\} in my \{Crop 3\}, with a density of \{Density 3\}. The temperature was \{Temperature 3\}, and the location was at \{Location 3\}. Could you please provide some control and management suggestions? \newline \textbf{Answer:} \{Non-chemical control solutions for Pest 3\} \newline \textbf{Question:} I discovered \{Pest\} in my \{Crop\}, with a density of \{Density\}. The temperature was \{Temperature\}, and the location was at \{Location\}. Could you please provide some control and management suggestions? \newline \textbf{Answer:}}

\subsubsection{Instruction-based Prompting}
\label{ib}
As mentioned in Section \ref{sec:prompt}, constructing an effective prompt can involve any of the four elements: Instruction, Context, Input Data and Output Indicator \cite{04}. Giray \cite{029} discussed the importance of understanding the prompt component and its role in facilitating effective communication with the model. Through prompt design with these four elements, Giray \cite{029} found one can guide model behaviour and improve response quality, ensuring output is precise and meaningful. The template of the instruction-based prompt is: \emph{\newline \textbf{Instruction:} Generate comprehensive and sustainable pest management suggestions based on the given crop, pest type and density, and environmental conditions, including temperature and location. \newline \textbf{Context:} Pest management in agriculture requires balancing control measures with environmental sustainability. Different crops and pests respond to varied strategies, and local environmental conditions significantly influence the effectiveness of these strategies. \newline \textbf{Input Data:} For example:
\newline For pest: \{Pest\}
\newline The affected crops are: \{Affected Crops\}
\newline The threshold is: \{Threshold\}
\newline The non-chemical control solution could be: \{Non-chemical control solutions\}
\newline \textbf{Output Indicator:} Question: I discovered \{Pest\} in my \{Crop\}, with a density of \{Density\}. The temperature was \{Temperature\}, and the location was at \{Location\}. Could you please provide some control and management suggestions? Please first determine whether management measures are needed, then output your own control solution in about 200 words.}

The Instruction defines the pest management task for the model and guides the model to focus on the data in the input question. The context explains why pest management in agriculture is essential, helping the model better understand the broader implications of pest management and the necessity of tailoring suggestions to specific scenarios. The structured example systematically introduces the input data, incorporating placeholders for designated variables. Precisely, the \emph{\{Pest\}} variable corresponds to the pest identified in the inquiry, while the \emph{\{Affected Crops\}} are derived from the `pest\_to\_affected\_crop.json' file. Similarly, the \emph{\{Threshold\}} values are extracted from the `thresholds\_database.xlsx' file, and the \emph{\{Non-chemical control solutions\}} are obtained from the `pest\_to\_management.xlsx' file. All input data variables are dynamically populated based on the specific pest mentioned in the question.

\subsubsection{Self-consistency Prompting}
\label{ss}
Self-consistency is an advanced prompting technique introduced by Wang et al. \cite{030} building upon chain-of-thoughts (CoT). This innovative approach involves generating diverse reasoning paths rather than relying on the most immediately probable path. Self-consistency then deduces the most consistent answer by aggregating across these varied reasoning paths. Self-consistency prompting summarised the responses from the zero-shot, few-shot, and instruction-based prompting and gave a final response. The template of self-consistency prompting is: \emph{\newline Given these three responses: \newline Response 1: \{Response 1 from zero-shot prompting\}\newline Response 2: \{Response 2 from few-shot prompting\}\newline Response 3: \{Response 3 from instruction-based prompting\}\newline Create a summary response that combines the best elements of question: I discovered \{Pest\} in my \{Crop\}, with a density of \{Density\}. The temperature was \{Temperature\}, and the location was at \{Location\}. Could you please provide some control and management suggestions?}

\subsection{GPT-4 as Evaluator}
\label{sec:evaluation}

Twelve combinations emerge when integrating FLAN, GPT-3.5, and GPT-4 models with four prompting methodologies. Each combination is subjected to fifty input pest samples characterized by varying density and environmental conditions, generating respective responses. These responses are then evaluated by GPT-4 (Model: `gpt-4-1106-preview') regarding the accuracy and the linguistic quality of the generated pest management suggestions. The prompt guiding the GPT to serve as an evaluator for determining the necessity of action in responses and assessing the linguistic quality of these responses draws inspiration from the article ``G-EVAL: NLG Evaluation using GPT-4 with Better Human Alignment'', where the article introduces the G-EVAL framework, designed for evaluating the quality of text generated by Natural Language Generation (NLG) systems \cite{031}.

For accuracy evaluation, the prompt begins with the `Evaluation Guide', which instructs the GPT to assess and decide whether action is required. The prompt followed with the `Evaluation Criteria' to inform the GPT that this is a binary evaluation to assign `1' or `0'. Through the Evaluation Steps, the GPT is guided with the CoT sequence, asked to decide whether the information presented within a response indicates that an action is required (`1') or not required (`0') in the following prompt and the corresponding response. The template for accuracy evaluation is shown below:
\emph{\newline \textbf{Evaluation Guide:} \newline You will be provided with a prompt and the corresponding response for pest management.\newline Your task is to evaluate the response based on the criteria below and decide whether action is required based on the response.\newline Please read and understand these instructions carefully. Refer back to this document as needed during your evaluation. \newline \textbf{Evaluation Criteria:} \newline 
Action Required (1 or 0) - This is a binary evaluation to determine if action is needed based on the response provided. \newline \textbf{Evaluation Steps:}\newline 1. Carefully read the pest management suggestion in the response, identifying the main content, pay special attention to the first sentence in the response, as it generally contains the decision of whether to take actions.\newline 2. Analyze the response to see if it states whether action is required or not required to manage the pest. \newline 3. Assign a score based on the evaluation criteria: 0 means no action is needed, 1 means the suggestion requires action. \newline 4. If the response suggests the action is optional, needs further observation or continuous monitoring, leaves room for doubt, lacks clearly direction, contains not be necessary or not immediate control, or if you cannot determine with complete certainty that it indicates for management action, please mark it as 0.\newline \textbf{Here are the prompt and response you need to evaluate:}\newline \textbf{Prompt:} \{Prompt\} \newline \textbf{Response:} \{Response\}\newline \textbf{Please state whether action is required (Answer 0 or 1 ONLY):}}

The linguistic quality evaluation contains six dimensions: Coherence, Logical Consistency, Fluency, Relevance, Comprehensiveness, and Exhaustiveness. The structure of the prompt for linguistic quality evaluation is similar to accuracy evaluation, comprising an Evaluation Guide with instructions, Evaluation Criteria that include scoring standards, and Evaluation Steps based on a CoT approach. Except for some differences in details and descriptors, the principal distinction lies in the judgment required from the GPT. For accuracy evaluation, the GPT is tasked with making a binary decision regarding the necessity of action. In contrast, evaluating linguistic quality required the GPT to assign scores ranging from 1 to 10 for each of the six dimensions.

\section{Results}
\label{sec:result}

\begin{table}[H]
\centering
\begin{tabular}{lcccccccc}
\hline
\textbf{Model \& Prompting} & \textbf{Coherence} & \textbf{Consistency} & \textbf{Fluency} & \textbf{Relevance} & \textbf{Comprehensibility} & \textbf{Exhaustiveness}  \\ \hline
 FLAN zero-shot              & 2.52               & 2.52                         & 3.30              & 2.36               & 2.76                      & 2.96                                   \\ 
FLAN few-shot               & 2.68               & 3.00                         & 3.42              & 2.44               & 3.32                      & 3.46                                  \\ 
 FLAN instruction-based      & 3.70               & 3.92                         & 4.84              & 5.06               & 5.04                      & 4.36                                   \\ 
 FLAN self-consistency       & 2.64               & 3.22                         & 4.04              & 1.94               & 3.92                      & 3.18                         
 \\ \hline
 
 GPT-3.5 zero-shot           & 8.82               & 8.24                         & 9.90              & 8.74               & 9.54                      & 7.54                                   \\ 
GPT-3.5 few-shot            & 8.14               & 8.24                         & 9.86              & 9.26               & 8.36                      & 6.28                                  \\ 
GPT-3.5 instruction-based   & 8.28               & 8.20                         & 9.60              & 8.92               & 9.14                      & 6.92                                   \\ 
GPT-3.5 self-consistency    & 7.98               & 8.00                         & 9.80              & 7.70               & 9.44                      & 7.16                                  \\ \hline
GPT-4 zero-shot             & 9.14               & 8.88                         & 10.00             & 9.86               & 9.38                      & 8.74                                  \\ 
 GPT-4 few-shot              & 8.32               & 8.46                         & 9.98              & 9.46               & 8.92                      & 7.14                                  \\ 
 GPT-4 instruction-based     & 8.62               & 8.76                         & 9.64              & 9.46               & 9.32                      & 7.68                                   \\ 
 GPT-4 self-consistency      & 8.72               & 8.90                         & 10.00             & 9.30               & 9.88                      & 8.14                                   \\ \hline
\end{tabular}
\caption{Linguistic quality of different models and prompting methods evaluated by GPT-4}
\label{table:model_comparison}
\end{table}

\begin{table}[H]
\centering
\begin{tabular}{lcccccccccc}
\hline
\textbf{Model \& Prompting} & \textbf{TP} & \textbf{TN} & \textbf{FP} & \textbf{FN} & \textbf{Accuracy} & \textbf{Precision} & \textbf{Recall} & \textbf{F1 Score} & \textbf{Final Score} \\ \hline
FLAN zero-shot              & 20 & 6  & 19 & 5  & \textbf{0.52} & 0.51 & 0.80 & 0.62 & \textbf{37.22} \\ 
FLAN few-shot               & 10 & 10 & 15 & 15 & \textbf{0.40} & 0.40 & 0.40 & 0.40 & \textbf{34.32} \\ 
FLAN instruction-based      & 14 & 14 & 11 & 11 & \textbf{0.56} & 0.56 & 0.56 & 0.56 & \textbf{49.32} \\ 
FLAN self-consistency       & 24 & 1  & 24 & 1  & \textbf{0.50} & 0.50 & 0.96 & 0.66 & \textbf{38.94} \\ \hline
GPT-3.5 zero-shot           & 25 & 4  & 21 & 0  & \textbf{0.58} & 0.54 & 1.00 & 0.70 & \textbf{75.98} \\ 
GPT-3.5 few-shot            & 17 & 8  & 17 & 8  & \textbf{0.50} & 0.50 & 0.68 & 0.58 & \textbf{70.14} \\ 
GPT-3.5 instruction-based   & 24 & 12 & 13 & 1  & \textbf{0.72} & 0.65 & 0.96 & 0.77 & \textbf{79.86} \\ 
GPT-3.5 self-consistency    & 25 & 0  & 25 & 0  & \textbf{0.50} & 0.50 & 1.00 & 0.67 & \textbf{70.08} \\ \hline
GPT-4 zero-shot             & 24 & 4  & 21 & 1  & \textbf{0.56} & 0.53 & 0.96 & 0.69 & \textbf{78.40} \\ 
GPT-4 few-shot              & 21 & 7  & 18 & 4  & \textbf{0.56} & 0.54 & 0.84 & 0.66 & \textbf{74.68} \\ 
GPT-4 instruction-based    & 24 & 9  & 16 & 1  & \textbf{0.66} & 0.60 & 0.96 & 0.74 & \textbf{79.88} \\ 
GPT-4 self-consistency     & 25 & 0  & 25 & 0  & \textbf{0.50} & 0.50 & 1.00 & 0.67 & \textbf{74.94} \\ \hline
\end{tabular}
\caption{Performance metrics of different models and prompting methods with final scores}
\label{table:performance_metrics}
\end{table}

Tables \ref{table:model_comparison} and \ref{table:performance_metrics} respectively present the linguistic quality of different models and prompting methods evaluated by GPT-4, and the performance metrics of different models and prompting methods with the final scores for each model. The linguistic quality evaluation involves scoring the responses based on the generated pest management suggestions across 50 samples, each representing an average derived from these responses. In performance metrics, the TP (True Positives), TN (True Negatives), FP (False Positives), and FN (False Negatives) are used as foundational elements for calculating Accuracy, Precision, and Recall.

To calculate the final scores for each ``Model \& Prompting'' combination, we use a weighted average approach based on pre-determined weights for various evaluation metrics. Specifically, the weights for Coherence, Logical Consistency, Fluency, Relevance, Comprehensibility, and Exhaustiveness are each allocated $10\%$, while Accuracy is assigned a higher weight of $40\%$. In the computation of the Final Score, the metrics of Coherence, Consistency, Fluency, Relevance, Comprehensibility, and Exhaustiveness are evaluated on a scale from 1 to 10. In contrast, Accuracy is averaged, falling in a range from 0 to 1. To harmonize these scores for a unified presentation in a percentage format, the scores for the linguistic quality are multiplied by $10$, and the Accuracy score is multiplied by $100$, facilitating a standardized evaluation outcome expressed on a 100-point scale. The mathematical formulation for the final score for each model can be expressed as follows:
\begin{equation}
\begin{aligned}
Final\ Score = & \, 0.1 \times (Val_{\text{Coherence}} + Val_{\text{Consistency}} + Val_{\text{Fluency}} + Val_{\text{Relevance}}\\
               &  + Val_{\text{Comprehensibility}} + Val_{\text{Exhaustiveness}}) \times 10 \\
               & + 0.4 \times Val_{\text{Accuracy}} \times 100
\end{aligned}
\end{equation}

Where $Val_{\text{Coherence}}$, $Val_{\text{Consistency}}$, $Val_{\text{Fluency}}$, $Val_{\text{Relevance}}$, $Val_{\text{Comprehensibility}}$ and $Val_{\text{Exhaustiveness}}$ respectively represent the numerical values scored by GPT-4 for the dimensions of Coherence, Consistency, Fluency, Relevance, Comprehensibility and Exhaustiveness as listed in Table \ref{table:model_comparison}, and $Val_{\text{Accuracy}}$ denotes the average accuracy value in Table \ref{table:performance_metrics}.

From Table \ref{table:model_comparison}, it can be observed that the performance of the different models and their application of different prompting methods on the various dimensions of language quality. Specifically, the FLAN model scores low on each assessed dimension, showing its understanding and generating language limitations. For example, FLAN zero-shot scored no more than 3.3 on cohesion, logical consistency, fluency, relevance, comprehensibility, and exhaustiveness, indicating that the FLAN model struggles to handle complex language tasks effectively without specific training or guidance in generating pest management suggestions. In contrast, the GPT-3.5 and GPT-4 models scored significantly higher than the FLAN model on all dimensions, especially GPT-4, which achieved a perfect score of 10 on fluency and scored higher than 8 on the remaining dimensions. This result also demonstrates the excellent ability of GPT-3.5 and GPT-4 to generate high-quality, logically consistent, and relevant suggestions in pest management. It is worth noting that the same model scores roughly the same on all dimensions of linguistic quality using different prompting methods, suggesting that variations in prompting method have a limited impact on the language quality of the model output. For example, the scores of the FLAN model under different prompting methods are different, but the overall performance is still poor. In contrast, the GPT-3.5 and GPT-4 models maintain a high level of performance on all dimensions regardless of the prompting method used.

Table \ref{table:performance_metrics} shows differences in performance metrics across models and prompting methods, focusing on accuracy, precision, recall, and F1 scores. Evidently, the GPT-3.5 and GPT-4 models outperform the FLAN model across nearly all metrics, indicating their superior ability to generate pest management advice. Interestingly, while most models exhibit high recall rates, their accuracy and precision remain low. This suggests that although the models can identify positive samples which require action in pest scenarios, they did the wrong classification in scenarios scenarios that not require an action, leading to a high rate of FP. Moreover, the performance impact of different prompting methods on the same model varies. For example, the accuracy of the instruction-based method outperforms other prompting methods for the same model. This is attributed to including pest threshold levels and affected crops in the instruction-based prompts, enabling LLMs to make better-informed judgments in pest management based on the information provided in the prompts.

The instruction-based method with GPT-3.5 demonstrates the best performance in accuracy, precision, recall, and F1 scores. Unexpectedly, it even surpasses GPT-4. Examination of model responses reveals that although GPT-4 may better understand the content of prompts or appear ``smarter", it occasionally makes judgments such as ``Although your current density is not at the advised threshold level, preventive measures should be taken before populations reach damaging thresholds" (indicating action despite not reaching the threshold) or ``Although ... are not typical pests of ..., they may occasionally be found on various crops" (classifying a non-affected pest as potentially affecting other crops, thereby suggesting that action is needed). This leads to GPT-4 inaccurately classifies negative samples. Meanwhile, GPT-3.5 adheres strictly to the thresholds specified in the prompts, more inclined to conclude that ``... does not currently reach the treatment threshold, management measures may not be immediately necessary'' so that making more accurate judgments on negative samples.

The self-consistency prompting exhibits the poorest performance among all prompting methods. Despite its ability to correctly identify almost all positive samples, it incorrectly classifies nearly all negative samples as positive, suggesting that self-consistency prompts the model to judge nearly every scenario as requiring action. This outcome is due to the prompt containing a directive to ``Create a summary response that combines the best elements", asking the model to summarize based on responses from zero-shot, few-shot, and instruction-based prompting. Given the model's inherent insensitivity to negative samples, such summarization further deteriorates precision.

The instruction-based scores of GPT-4 are comparable to those of GPT-3.5, both around 79, indicating a clear advantage in pest management scenarios when the model is provided with affected crops and threshold information in the prompt. In contrast, FLAN model scores are generally lower, with its instruction-based score reaching 49.32 but still below those of GPT series models, reflecting FLAN's limitations in agricultural domain knowledge. The self-consistency prompting method performs relatively better in GPT-3.5 and GPT-4 models but still scores below instruction-based prompting due to its tendency to classify nearly all negative scenarios as positive. Zero-shot and few-shot methods score lower across all models, likely due to their lack of sufficient contextual information to guide the model in generating the most relevant and accurate advice.

\section{Conclusion}
\label{sec:conc}

In conclusion, this study evaluated the ability of different LLMs to generate suggestions for pest management in agriculture using different prompting methods. By simulating various pest scenarios, we understood the strengths and limitations of using different LLMs and prompting methods, bridging the gap between the lack of research on LLMs in agriculture. GPT-3.5 and GPT-4 showed accuracy and relevance in delivering pest management solutions, demonstrating the potential of GPT-4 as an agricultural support tool. However, LLMs tended to generate generic suggestions and showed less sensitivity in facing negative samples. This highlights the need for continuous model updating and domain-specific fine-tuning. Instruction-based prompting led to a significant increase in the accuracy of LLMs, confirming that the addition of relevant knowledge domains has an indispensable role in generating responses to LLMs. In the future, we aim to enhance prompting methodologies to enable LLMs to generate more precise evaluations by integrating domain-specific knowledge. Simultaneously, by refining the prompts, we aspire for LLMs to deliver more detailed and user-friendly responses. Given that this technology primarily targets farmers, it is advantageous if responses can provide differentiated pest control methods tailored to various pest stages, including recommendations on varying dosages for management, intervals for prevention and control or subsequent monitoring.

\bibliographystyle{IEEEbib}
\bibliography{refs.bib}

\end{document}